\documentclass[10pt,twocolumn,letterpaper]{article}
\usepackage[pagenumbers]{cvpr}
\usepackage{amsmath,amssymb,amsthm,mathtools}
\usepackage{graphicx,booktabs,multirow,tabularx,longtable}
\usepackage{algorithm,algpseudocode}
\usepackage{microtype}
\usepackage{needspace}
\definecolor{cvprblue}{rgb}{0.21,0.49,0.74}
\usepackage[pagebackref,breaklinks,colorlinks,allcolors=cvprblue]{hyperref}

\newcommand{\D}{\mathcal{D}}

\newcommand{\calL}{\mathcal{L}}
\newcommand{\norm}[1]{\left\lVert#1\right\rVert}



\DeclareMathOperator*{\argmin}{arg\,min}
\hypersetup{pdftitle={Calibrate in Context: Efficient Data-Free Quantization from Small Synthetic Pools}}

\usepackage{placeins}
\usepackage{float}
\title{Calibrate in Context:\\Efficient Data-Free Quantization from Small Synthetic Pools}
\makeatletter
\patchcmd{\@maketitle}{\vspace*{24pt}}{\vspace*{6pt}}{}{\PackageError{fgcpreprint}{Title spacing patch failed}{}}
\patchcmd{\@maketitle}{\vskip .375in}{\vskip .08in}{}{\PackageError{fgcpreprint}{Title top spacing patch failed}{}}
\patchcmd{\@maketitle}{\vspace*{12pt}}{\vspace*{2pt}}{}{\PackageError{fgcpreprint}{Title bottom spacing patch failed}{}}
\makeatother
\author{%
Haowen Hua\textsuperscript{1}\quad Sen Lin\textsuperscript{1}\quad Yaokun Huang\textsuperscript{1}\quad Zeyi Shao\textsuperscript{1}\quad Hao Zhang\textsuperscript{1}\\[1pt]
Wei Sui\textsuperscript{2}\quad John See\textsuperscript{3}\quad Zeyd Boukhers\textsuperscript{4}\quad Cong Yang\textsuperscript{1,*}\\[2pt]
{\small\textsuperscript{1}Soochow University\quad\textsuperscript{2}D-Robotics\quad\textsuperscript{3}Heriot-Watt University}\\[1pt]
{\small\textsuperscript{4}Fraunhofer Institute for Applied Information Technology}\\[1pt]
{\small\textsuperscript{*}Corresponding author: \href{mailto:cong.yang@suda.edu.cn}{\texttt{cong.yang@suda.edu.cn}}}%
}
\hypersetup{pdfauthor={Haowen Hua, Sen Lin, Yaokun Huang, Zeyi Shao, Hao Zhang, Wei Sui, John See, Zeyd Boukhers, Cong Yang}}

\begin{document}
\maketitle
\begin{abstract}
We enable accurate low-bit recovery from small synthetic pools by improving both image synthesis and student preparation. A grid with lower local response error can incur greater error in the student's block context. Functional Grid Calibration (FGC) selects activation scales and integer offsets against full-precision responses constructed from the student's current block input. With weights fixed during calibration, FGC evaluates candidates only at their consuming operators, bringing block context to calibration at nearly local-scoring cost. For DeiT and ViT, progressive inversion updates the same images through increasingly deep teacher prefixes. With 1,000 images, the complete pipeline exceeds MimiQ's reported accuracy in all 18 W4A4/W5A5 settings across nine backbones. With 256 images, DeiT-S and ViT-S reach 68.20\% and 66.90\% ImageNet top-1 accuracy at W4A4 in 26--27 minutes on one RTX 5090. The pipeline outperforms strengthened 1K-image MimiQ by 13.03 and 24.36 percentage points at lower preparation cost and the same recovery-update count. The same low-budget checkpoints retain their advantage on ImageNetV2.
\end{abstract}

\section{Introduction}\label{sec:intro}
Data-free quantization prepares low-bit students without the original training images. Its preparation cost includes both synthesizing inputs and recovering the student. We make small synthetic pools effective by improving how they are generated and how the student is prepared to use them.

Existing methods improve synthetic inputs through patch relations, inter-head structure, semantics and informative regions~\cite{psaq,psaqv2,mimiq,sardfq,maskaq,clampvit}. The usefulness of these inputs also depends on the student's activation grids: scales and integer offsets determine clipping and code assignments, shaping the errors from which recovery starts.

\begin{figure*}[t]
\centering\includegraphics[width=\textwidth]{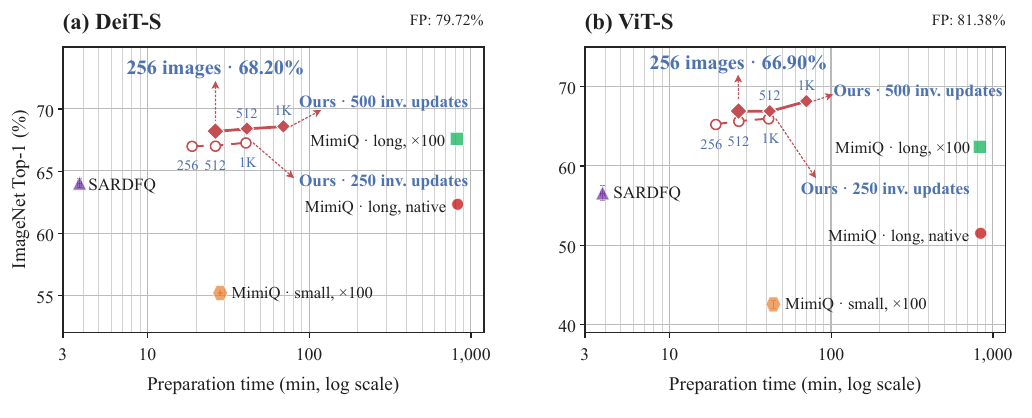}
\caption{\textbf{High accuracy from small synthetic pools at low preparation cost.} Our 256-image configurations outperform reduced-budget MimiQ at lower total preparation cost. W4A4 timings include synthesis, calibration and recovery on one RTX 5090. Labels indicate pool size; each curve fixes the inversion updates per image. Our bars span two recovery seeds. $100\times$ denotes amplified ordinary-activation scale gradients; external configurations are defined in \cref{sec:protocol}.}\label{fig:external}
\end{figure*}

A locally more accurate grid can be worse in context. An operator receives inputs distorted by earlier quantization. A local target applies the full-precision (FP) operator to the current student operands, which already contain within-block distortion. Reducing this local error can remove useful cancellation with preceding within-block distortion and increase contextual error (\cref{fig:reference-mechanism}).

We turn this observation into \emph{Functional Grid Calibration} (FGC). Starting at the student's current block input, FGC follows an FP path and caches its operator responses. This matches the student's entry state; a full-teacher reference follows the complete FP trajectory from the input image. With weights fixed, FGC evaluates candidate activation grids on current student operands, rerunning only the consuming operator. FGC thus uses block context to choose grids while keeping candidate evaluation local.

For flat transformers, progressive inversion updates the same images through increasingly deep teacher prefixes, retaining optimizer states. Both synthesis and calibration improve recovery with the other held fixed (\cref{tab:components}a). At near-equal preparation cost, longer inversion of 256 images also outperforms shorter inversion of 512 on four DeiT/ViT small and base models.

Our contributions are:
\begin{itemize}
\item \textbf{Contextual calibration with local evaluation.} FGC constructs targets from the student's block input and searches activation grids at their consuming operators, capturing useful error interactions at nearly local-scoring cost.
\item \textbf{Progressive allocation of synthesis work.} Progressive inversion improves pool utility at matched logical work and, separately, matched image-update counts. Budget studies show the value of longer inversion of fewer images.
\item \textbf{Accurate recovery from small synthetic pools.} The pipeline delivers accurate W4A4/W5A5 models across nine backbones. DeiT-S/ViT-S recover from 256 images in 26--27 minutes and retain their advantage on ImageNetV2.
\end{itemize}

\section{Related work}
\paragraph{Synthetic data for quantization.}
DeepInversion synthesizes inputs from teacher statistics~\cite{deepinversion}. For transformers, PSAQ-ViT exploits patch similarity, and PSAQ-ViT V2 couples synthesis with student optimization~\cite{psaq,psaqv2}. MimiQ introduces inter-head relations and attention distillation~\cite{mimiq}; SARDFQ adds semantic alignment and local enrichment~\cite{sardfq}; MaskAQ emphasizes informative regions and refreshes synthetic samples~\cite{maskaq}. CLAMP-ViT alternates contrastive synthesis with layer-wise quantization search~\cite{clampvit}. QuaSC uses activation correction without fine-tuning~\cite{dfqvit}, while diffusion-assisted methods use external generative priors~\cite{mixupprompt,jointptq}. We allocate teacher computation across depths and prepare the activation grids through which the synthetic inputs train the student.

\paragraph{Functional calibration and reconstruction.}
LSQ and LSQ+ establish learned step sizes and mean-squared-error (MSE) initialization~\cite{lsq,lsqplus}. EasyQuant alternates weight- and activation-scale search using quantized-prefix inputs and original-model output targets~\cite{easyquant}, and APQ-ViT uses block calibration with specialized probability quantization~\cite{apqvit}. AdaRound introduces asymmetric reconstruction~\cite{adaround}; GPTAQ uses FP-trajectory targets to account for accumulated error during weight reconstruction~\cite{gptaq}. BRECQ and QDrop improve block reconstruction~\cite{brecq,qdrop}, while FIMA-Q and LS-ViT refine its metric~\cite{fima,lsvit}. FGC restarts the FP reference at the student's block input and scores activation-grid candidates only at their consuming operators, with pretrained weights fixed.

\paragraph{Efficient and stable preparation.}
Initialization sensitivity and quantization oscillation motivate specialized training procedures~\cite{lsqplus,oscillations,ofq}. Sparse Model Inversion reduces synthesis work through spatial selection~\cite{smi}, and Patch Rebirth detaches informative patches while continuing to invert the remainder~\cite{pri}. Progressive synthesis allocates updates across teacher depths while retaining all tokens.

\section{Efficient data-free preparation}\label{sec:method}
\begin{figure*}[t]
\centering\includegraphics[width=\textwidth]{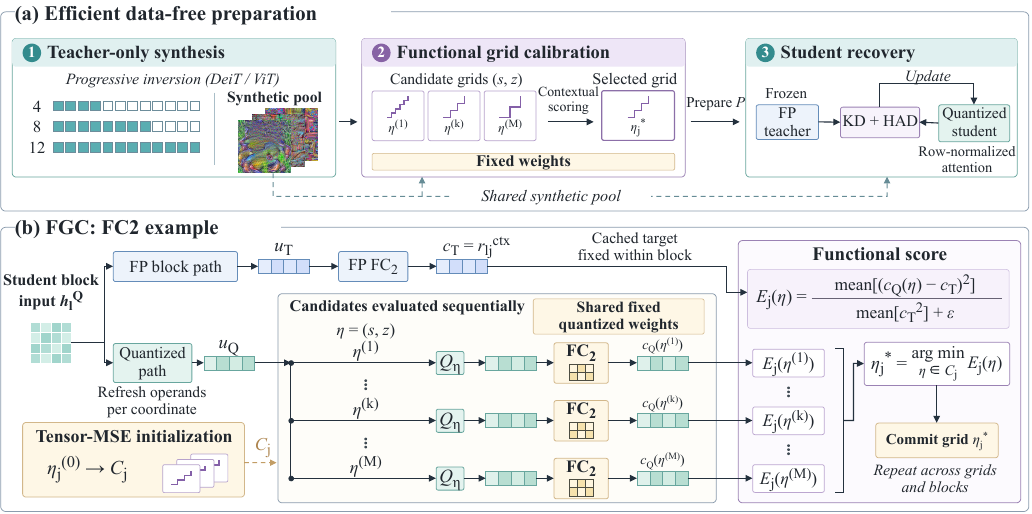}
\caption{\textbf{Generating a small pool and preparing the student to use it.} (a) The same synthetic pool supplies calibration and recovery views. DeiT/ViT use progressive inversion; quantitative Swin results use full-depth pools. The final probability representation is installed after calibration, with row normalization active during recovery and inference. (b) Stage 2 illustrated at FC2, the MLP output projection. Both paths start from the same student block input. The cached FP target stays fixed within the block. Candidates are evaluated sequentially with shared current operands and fixed quantized weights; operands are refreshed before each coordinate search.}
\label{fig:method}
\end{figure*}

Our pipeline generates a synthetic pool, prepares activation grids with FGC, and recovers the student by distillation (\cref{fig:method}a). FGC chooses grids by comparing student operator responses with cached FP responses from the same student block input (\cref{fig:method}b).

\subsection{Constructing the reference in student context}
Let $h_l^Q(x)$ be the input to block $l$ produced by the current quantized prefix. Both the FP block and student start from this state. FGC caches the FP operator responses along the block; the student follows its quantized path, producing potentially different internal operands.

Let $u_{lj}^Q$ and $u_{lj}^T$ denote the pre-quantization student and FP operands at the operator consuming grid $j$. Matrix products use operand tuples, with one grid coordinate varied at a time. FGC uses $r_{lj}^{\rm ctx}=C_{lj}^T(u_{lj}^T)$, the FP response along the within-block path from $h_l^Q$. Scoring against this target accounts for within-block discrepancies; candidate evaluation stops at the consuming operator.

At the same operator tap, the \emph{local} reference $r_{lj}^{\rm loc}=C_{lj}^T(u_{lj}^Q)$ applies the FP operator to student operands; the \emph{full-teacher} reference follows the complete FP trajectory from image $x$.

\subsection{Operator readout and error interaction}
\begingroup
\postdisplaypenalty=10000
Each activation grid maps its input to the affine quantization levels
\begin{equation}
 Q_{s,z}(x)=s\!\left[\operatorname{round}\!\left(\operatorname{clip}_{q_-,q_+}(x/s+z)\right)-z\right].
 \label{eq:grid}
\end{equation}
Here $s>0$, $z$ is an integer offset, and signed or unsigned code bounds are fixed. Neighboring offsets around an in-range incumbent obey $z\in[q_-,q_+]\cap\mathbb Z$, preserving exact zero. Attention-probability (P) grids use $z=0$.
\par\endgroup

For candidate $\eta_j=(s_j,z_j)$, $C_{lj}^Q$ quantizes the inputs and executes the consuming operator with fixed quantized weights where present. Let $M_{lj}$ count all scalar outputs over the 64 scoring views $\mathcal B_{\rm score}$. The normalized score is
\begin{equation}
 E_{lj}^{\rm op}(\eta_j)=
 \frac{M_{lj}^{-1}\!\sum_{x\in\mathcal B_{\rm score}}\norm{C_{lj}^{Q}(u_{lj}^{Q}(x);\eta_j)-r_{lj}^{\rm ctx}(x)}_F^{2}}
 {M_{lj}^{-1}\!\sum_{x\in\mathcal B_{\rm score}}\norm{r_{lj}^{\rm ctx}(x)}_F^{2}+\epsilon}.
 \label{eq:consumer}
\end{equation}
Here $\epsilon=10^{-12}$ and energies are pooled across the scoring bank. Cached operands and FP responses make candidate evaluation operator-local; the block-readout ablation scores the complete unit output.

Context changes the score through error interaction. At a fixed coordinate state, suppress indices and define
\begin{equation}
 e_\eta=C_Q(u_Q;\eta)-r^{\rm loc},\qquad
 b=r^{\rm loc}-r^{\rm ctx}.
 \label{eq:reference-gap}
\end{equation}
The contextual error decomposes exactly as
\begin{equation}
 \norm{C_Q(u_Q;\eta)-r^{\rm ctx}}^2
 =\norm{e_\eta}^2+2\langle e_\eta,b\rangle+\norm{b}^2.
 \label{eq:reference-score}
\end{equation}
The candidate error $e_\eta$ includes activation, fixed-weight and other fixed operator errors; $b$ is the fixed operator-output discrepancy induced by the two within-block paths. Local scoring uses $\norm{e_\eta}^2$; FGC also includes $2\langle e_\eta,b\rangle$. Cancellation with $b$ can make a larger local error yield a lower contextual score (\cref{fig:reference-mechanism}). FGC applies the asymmetric reconstruction principle~\cite{adaround,gptaq} to a student-block reference and an operator-level grid search.

\subsection{Grid proposals and sequential search}
The search has two stages: tensor-MSE initialization and contextual refinement. We construct a fixed bank from 128 synthetic images, each with one unaugmented and three recovery-augmented views. From a shared quantized state, we collect sampled activation values $\mathcal V_i$ for quantizer $i$ and select
\begin{equation}
 \eta_i^{(0)}\in\argmin_{(s,z)\in\mathcal C_i^{\rm init}}
 \sum_{u\in\mathcal V_i}\norm{Q_{s,z}(u)-u}^{2}.
 \label{eq:proposal}
\end{equation}
Candidates include the incumbent, observed-range and logarithmically spaced scales, and legal integer offsets; probability grids use unsigned support and zero offset. Proposals are committed after collecting all statistics.

The functional sweep uses 64 views from 64 distinct base images, balanced across the four view sets. At each ordinary coordinate, FGC tests at most nine scale--offset pairs: three scale multipliers $\{0.875,1,1.125\}$ and adjacent legal integer offsets. The incumbent is retained unless another candidate lowers the score. Matrix-product inputs are separate coordinates. Student operands are refreshed before each coordinate search and held fixed across its candidates. FP operator targets remain fixed within the unit. The prepared unit propagates its output to the next one; Swin includes patch merging. See Appendix~\ref{app:grid} for coordinate order and sampling caps.

\Cref{alg:fgcc} summarizes grid preparation and recovery.
\begin{algorithm}[t]
\caption{FGC grid preparation and student recovery}
\label{alg:fgcc}
\begin{algorithmic}[1]\small
\Require Teacher $T$, quantized student $S$, synthetic pool $\D$
\State Construct views $\mathcal V$ from $\D$; fix score indices $I$
\State Collect tensor samples; fit and commit MSE grids
\State $h\gets\mathrm{Prefix}_{S}(\mathcal V)$
\For{each block, merging, or classifier unit $l$}
\For{each distinct consuming-operator tap}
\State Run $T_l(h[I])$; cache that tap's FP response
\EndFor
\For{grid coordinate $j$ in the fixed order}
\State Run $S_l(h[I])$; cache pre-quantizer operands
\State Score hard candidates with Eq.~\eqref{eq:consumer}
\State Commit the best candidate; keep incumbent on ties
\EndFor
\State $h\gets S_l(h)$ \Comment{advance the full input bank}
\EndFor
\State Reset P grids; install row-normalized P operators
\State Freeze ordinary offsets; recover with KD and HAD
\State \Return recovered student $S$
\end{algorithmic}
\end{algorithm}

\subsection{Progressive allocation of synthesis work}
We spend early inversion updates on shallow teacher prefixes and increase depth while retaining the images and optimizer state. Let $P_d$ denote the depth-$d$ prefix. The readout $H_T$ is the teacher's existing final normalization and classification head. For a requested class $y$, the objective is
\begin{equation}
 \begin{aligned}
 \calL_d(x,y)={}&\mathrm{CE}(H_T(P_d(a_t(x))),y)\\
 &+\frac1L\sum_{l=1}^d\ell_{\mathrm{IHC},l}(a_t(x))
 +\beta\ell_{\mathrm{TV}}(x).
 \end{aligned}
 \label{eq:pg}
\end{equation}
Here $L$ is the teacher's full block count, $a_t$ the sampled input transformation, and $\beta=2.5\times10^{-5}$. The inter-head correlation (IHC) and image total-variation (TV) penalties, together with the image transformations, follow MimiQ~\cite{mimiq}; the IHC term retains full-depth normalization $1/L$, so the active inter-head terms expand with the prefix.

For 12-block DeiT/ViT teachers, PG500 uses depths 4/8/12 for 167/167/166 updates, respectively. Its logical block work equals 333 full-depth updates (Full333):
\[
167(4+8)+166\cdot12=3996=333\cdot12.
\]
Full500 uses 500 full-depth updates, or 6,000 block evaluations per image. \Cref{fig:synthesis-utility} (right) shows the schedules; we measure complete preparation time separately.

\subsection{Probability representation and recovery}
Quantized attention rows need not sum to one. We normalize each quantized row before value aggregation and initialize its unsigned grid with $z_P=0$ and scale
\begin{equation}
 s_P=\frac{2}{n_{\rm key}\sqrt{2^{b_P}-1}},
\label{eq:p-init}
\end{equation}
where $n_{\rm key}$ is the attention-row length and $b_P$ the probability bitwidth. This is the LSQ initialization $2\langle|p|\rangle/\sqrt{2^{b_P}-1}$~\cite{lsq} specialized to a nonnegative, unit-sum attention row, for which $\langle|p|\rangle=1/n_{\rm key}$.

For $q=Q_{s_P,0}(p)$ and $m=\sum_iq_i$,
\begin{equation}
 \bar p=\begin{cases}q/m,&m>0,\\ \mathbf1_{n_{\rm key}}/n_{\rm key},&m=0,\end{cases}
 \qquad o=\bar p\,Q(V).
 \label{eq:renorm}
\end{equation}
For shifted-window attention, the zero-row fallback is uniform over the legal unmasked keys. This format has $b_P$-bit numerators and an input-dependent denominator; row summation and division use floating arithmetic.

FGC searches temporary P grids with unnormalized quantized probabilities. After the sweep, we reset their scales using \cref{eq:p-init}, fix their offsets at zero, and install row normalization. The selected ordinary activation grids initialize recovery, with their integer offsets frozen.

Recovery follows MimiQ with output knowledge distillation (KD) and head-wise attention distillation (HAD), updating shadow weights, activation scales, and the original biases, normalization and embedding parameters. The teacher supplies targets on each augmented view; synthesis requests supply no hard recovery labels.

\section{Experiments}\label{sec:experiments}
We show how synthesis and calibration yield accurate, low-cost models, then examine the reference choices and recovery behavior behind these gains.
\begin{table*}[t]
\centering\footnotesize
\caption{\textbf{Data-free ImageNet Top-1 accuracy (\%) across nine backbones.} Transformer-specific methods are shown here; \cref{tab:app-headline} gives the complete historical comparison. $\ddagger$: results for prior methods reported by MimiQ~\cite{mimiq}; other external cells are author-reported, except $\dagger$, our SARDFQ Swin-T adaptation. Our complete pipeline uses 1K images and reports the mean over two recovery seeds; FP identifies our teachers. Bold/underline mark the highest/second-highest reported value per column and bitwidth, excluding FP. A dash denotes an unreported setting. Quantizer, synthesis, and recovery protocols are method-specific.}
\label{tab:headline}
\begin{tabular*}{\textwidth}{@{\extracolsep{\fill}}llrrrrrrrrr@{}}\toprule
 & & \multicolumn{3}{c}{ViT} & \multicolumn{3}{c}{DeiT} & \multicolumn{3}{c}{Swin}\\
\cmidrule(lr){3-5}\cmidrule(lr){6-8}\cmidrule(lr){9-11}
W/A & Method & T & S & B & T & S & B & T & S & B\\\midrule
FP & Our FP teacher & 75.34 & 81.38 & 84.41 & 72.02 & 79.72 & 81.74 & 80.92 & 82.90 & 84.75\\\midrule
\addlinespace[2pt]
4/4 & PSAQ-ViT$^{\ddagger}$ & 0.67 & 0.15 & 0.94 & 19.61 & 5.90 & 8.74 & 22.71 & 9.26 & 23.69\\
 & PSAQ-ViT V2$^{\ddagger}$ & 1.54 & 4.14 & 2.83 & 22.82 & 32.57 & 45.81 & 50.42 & 39.10 & 39.26\\
\addlinespace[2pt]
 & MimiQ & 42.99 & \underline{55.69} & 62.91 & 52.03 & 62.72 & 74.10 & 69.33 & 70.46 & 73.49\\
 & SARDFQ & -- & 50.32 & 51.84 & 52.06 & 62.29 & 72.17 & 68.74$^{\dagger}$ & \underline{74.74} & \underline{76.42}\\
 & MaskAQ & \textbf{44.90} & -- & \underline{70.53} & \underline{53.10} & 63.81 & \underline{74.41} & \underline{70.40} & -- & --\\
 & CLAMP-ViT & -- & -- & -- & -- & \textbf{69.01} & -- & -- & -- & --\\
 & Ours & \underline{44.66} & \textbf{67.89} & \textbf{71.26} & \textbf{58.62} & \underline{68.91} & \textbf{76.48} & \textbf{74.98} & \textbf{78.67} & \textbf{79.59}\\
\midrule
\addlinespace[2pt]
5/5 & PSAQ-ViT$^{\ddagger}$ & 17.66 & 23.37 & 16.80 & 53.36 & 47.35 & 57.23 & 58.63 & 76.33 & 57.80\\
 & PSAQ-ViT V2$^{\ddagger}$ & 40.21 & 63.59 & 74.29 & 55.18 & 65.30 & 73.16 & 69.77 & 80.55 & 79.80\\
\addlinespace[2pt]
 & MimiQ & 62.40 & 70.02 & 78.09 & 63.40 & 72.59 & 78.20 & 76.39 & \underline{80.75} & \underline{82.05}\\
 & SARDFQ & -- & \underline{74.31} & 70.70 & \underline{66.41} & \underline{74.06} & 78.72 & 75.81$^{\dagger}$ & 79.56 & 80.82\\
 & MaskAQ & \underline{63.01} & -- & \underline{78.60} & 64.27 & 73.18 & \underline{78.89} & \underline{76.82} & -- & --\\
 & Ours & \textbf{65.91} & \textbf{76.03} & \textbf{80.50} & \textbf{67.50} & \textbf{76.45} & \textbf{79.90} & \textbf{78.73} & \textbf{81.29} & \textbf{82.13}\\
\bottomrule\end{tabular*}

\end{table*}

\subsection{Protocol}\label{sec:protocol}
We evaluate tiny, small and base DeiT/ViT and Swin models~\cite{deit,vit,swin} on ImageNet~\cite{imagenet}. Each teacher supplies 1,000 synthetic images by default; preparation uses no real images or external generator. DeiT/ViT use progressive synthesis and Swin uses full-depth synthesis. Our default pipeline uses a common FGC score and candidate policy, followed by 12,474 SGD updates with temperature-5 KD, HAD coefficient 10 and strong augmentation. Seeds 17/29 vary recovery sampling and augmentation on each pool. Each bitwidth starts from a fresh representation, and we report the final checkpoint. Within each calibration comparison, variants share the synthetic pool, final probability representation, ordinary-offset policy and recovery schedule unless explicitly varied.

W/A/P denote weight, ordinary-activation and probability bitwidths. P equals A for our method and ablations; external methods retain their reported scope. Patch merging and the classifier are quantized. LayerNorm, Softmax computation, nonlinearities and residual arithmetic retain their original precision; probability normalization uses a floating-point denominator. We used ImageNet-50K during development and evaluate frozen checkpoints on ImageNetV2. Appendix~\ref{app:protocol} details data access and evaluation. Budget pools differ from the architecture comparison in \cref{tab:headline}.

\paragraph{Timing and external configurations.}
Time includes synthesis, calibration and recovery on the same RTX 5090; process startup and final evaluation are excluded, and shared generation is fully charged per recovered model. Small-budget MimiQ uses 1K images, Full500 synthesis and 12,474 recovery updates with a complete learning-rate schedule. Long-budget MimiQ uses 10K images, 2K inversion updates and 123,750 recovery updates (seed 17). The strengthened variants use $100\times$ ordinary-scale gradients and an execution cache for fixed quantizer-control flags (Appendix~\ref{app:external}). SARDFQ uses 32 images, 1,000 generation updates, reparameterization off and 100 reconstruction updates per unit per stage (two complete runs)~\cite{sardfq}.

\begin{table*}[t]\centering\small
\caption{\textbf{Both synthesis and student preparation improve recovery.} W4A4 ImageNet Top-1 means (\%) over two recovery seeds. (a) Full333/PG500 $\times$ MSE/FGC on paired 1K pools at matched logical block evaluations per image; Full500 is the separate equal-update reference. (b) Row normalization toggled from the same MSE state and mean-P scales on a separate 1K pool. (c) Tensor MSE versus FGC on the 256-image PG500 pools of \cref{fig:external}. Recovery follows the common protocol in \cref{sec:protocol}.}\label{tab:components}
\begin{minipage}[t]{.49\textwidth}\centering
\renewcommand{\arraystretch}{1.15}
\textit{(a) Synthesis $\times$ calibration, 1K}\\[3pt]
\begin{tabular*}{\linewidth}{@{\extracolsep{\fill}}llrrr@{}}\toprule
Synthesis & Calibration & Block evals. & DeiT-S & ViT-S\\\midrule
Full333 & Tensor MSE & 3,996 & 64.51 & 65.95\\
Full333 & FGC & 3,996 & 66.46 & 66.58\\
PG500 & Tensor MSE & 3,996 & 67.02 & 66.95\\
\textbf{PG500} & \textbf{FGC} & \textbf{3,996} & \textbf{68.43} & \textbf{67.68}\\\midrule
Full500 & FGC & 6,000 & 67.42 & 66.96\\\bottomrule
\end{tabular*}\end{minipage}\hfill
\begin{minipage}[t]{.22\textwidth}\centering
\renewcommand{\arraystretch}{1.38}
\textit{(b) Row normalization}\\[3pt]
\begin{tabular*}{\linewidth}{@{\extracolsep{\fill}}lcr@{}}\toprule
Model & Norm. & Top-1\\\midrule
\multirow{2}{*}{DeiT-S} & Off & 66.26\\
 & On & \textbf{66.81}\\\midrule
\multirow{2}{*}{ViT-S} & Off & 63.70\\
 & On & \textbf{66.95}\\\bottomrule
\end{tabular*}\end{minipage}\hfill
\begin{minipage}[t]{.25\textwidth}\centering
\renewcommand{\arraystretch}{1.38}
\textit{(c) Fig.~1 pools, 256 images}\\[3pt]
\begin{tabular*}{\linewidth}{@{\extracolsep{\fill}}llr@{}}\toprule
Model & Grid & Top-1\\\midrule
\multirow{2}{*}{DeiT-S} & MSE & 66.93\\
 & FGC & \textbf{68.20}\\\midrule
\multirow{2}{*}{ViT-S} & MSE & 66.27\\
 & FGC & \textbf{66.90}\\\bottomrule
\end{tabular*}\end{minipage}

\end{table*}

\subsection{Accuracy and complete preparation cost}\label{sec:accuracy-cost}
\paragraph{High accuracy with 256 synthetic images.}
Our pipeline reaches 68.20\%/66.90\% on DeiT-S/ViT-S in 26.36/26.80 minutes using 256 images (\cref{fig:external}). Strengthened 1K-image MimiQ reaches 55.17\%/42.54\% in 28.20/43.95 minutes under the same 12,474-update recovery budget. Our models improve accuracy by 13.03/24.36 points with 6.5\%/39.0\% less preparation time. FGC also improves recovery on these same 256-image pools (\cref{tab:components}c).

\paragraph{High accuracy across architectures.}
With 1K images, our pipeline has the highest listed accuracy in 16 of 18 W4A4/W5A5 settings and exceeds MimiQ's reported result in all 18 (\cref{tab:headline}). Four-bit DeiT-S, ViT-S and Swin-T reach 68.91\%, 67.89\% and 74.98\%.

\begin{figure*}[t]\centering\includegraphics[width=\textwidth]{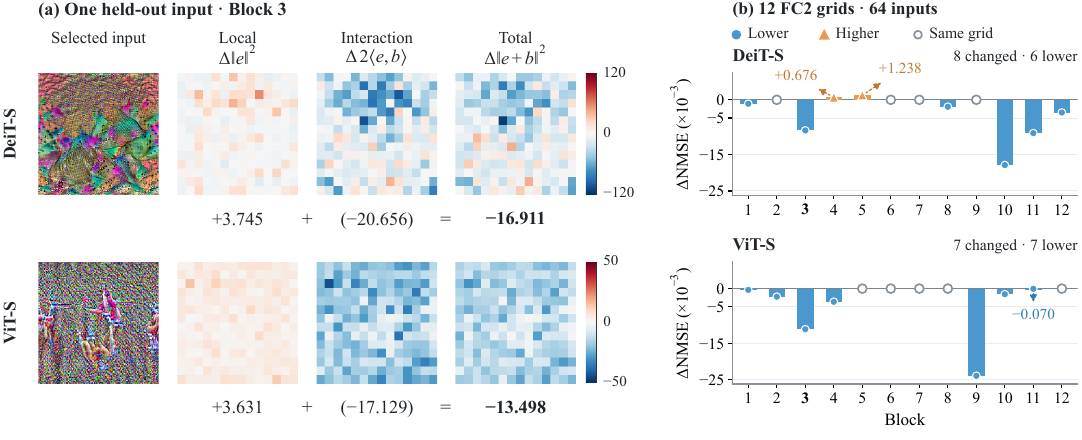}
\caption{\textbf{Context changes which grid is preferred.} (a) The contextual choice increases local operator-response error but improves the interaction enough to lower total error. (b) All 12 FC2 coordinates, including unchanged choices. Selection uses calibration views; evaluation uses 64 held-out synthetic inputs before recovery. Maps show individual inputs, with labels averaged over patch tokens; right panels pool all inputs and include CLS. Differences (contextual choice minus local choice) use contextual-reference normalization, in $10^{-3}$ units.}\label{fig:reference-mechanism}
\end{figure*}

\paragraph{Higher accuracy at much lower preparation cost.}
Our 1K-image DeiT-S/ViT-S configurations reach 68.59\%/68.15\% in 69.34/70.44 minutes (\cref{fig:external}), gaining 1.02/5.76 points over strengthened long-budget MimiQ in about one-twelfth the time. SARDFQ reaches 64.03\%/56.57\% in 3.80/3.86 minutes.

\begin{table*}[t]\centering\small
\caption{\textbf{Reference origin and scoring location in FGC.} Two-seed ImageNet Top-1 means (\%) with default recovery. (a) Reference origin at fixed operator readout. (b) Candidate readout with the student-block reference. Tensor MSE omits functional refinement. The panels use distinct pools.}\label{tab:calibration}
\begin{minipage}[t]{.49\textwidth}\centering
\textit{(a) Reference origin: operator readout fixed}\\[3pt]
\begin{tabular}{lrr}\toprule
Reference / preparation & DeiT-S & ViT-S\\\midrule
Tensor MSE &66.81&66.95\\
Local operator &67.87&67.39\\
Full-teacher trajectory &67.65&62.55\\
\textbf{Student block (FGC)} &\textbf{68.59}&\textbf{68.15}\\\bottomrule
\end{tabular}
\end{minipage}\hfill
\begin{minipage}[t]{.49\textwidth}\centering
\textit{(b) Evaluation scope: student-block reference fixed}\\[3pt]
\begin{tabular}{lrrr}\toprule
Readout / preparation & DeiT-S & ViT-S & Swin-T\\\midrule
Tensor MSE &66.99&67.30&74.32\\
Block &68.62&67.83&74.82\\
\textbf{Operator (FGC)} &\textbf{68.91}&\textbf{67.89}&\textbf{74.98}\\\bottomrule
\end{tabular}
\end{minipage}

\end{table*}

\begin{figure*}[t]\centering\includegraphics[width=\textwidth]{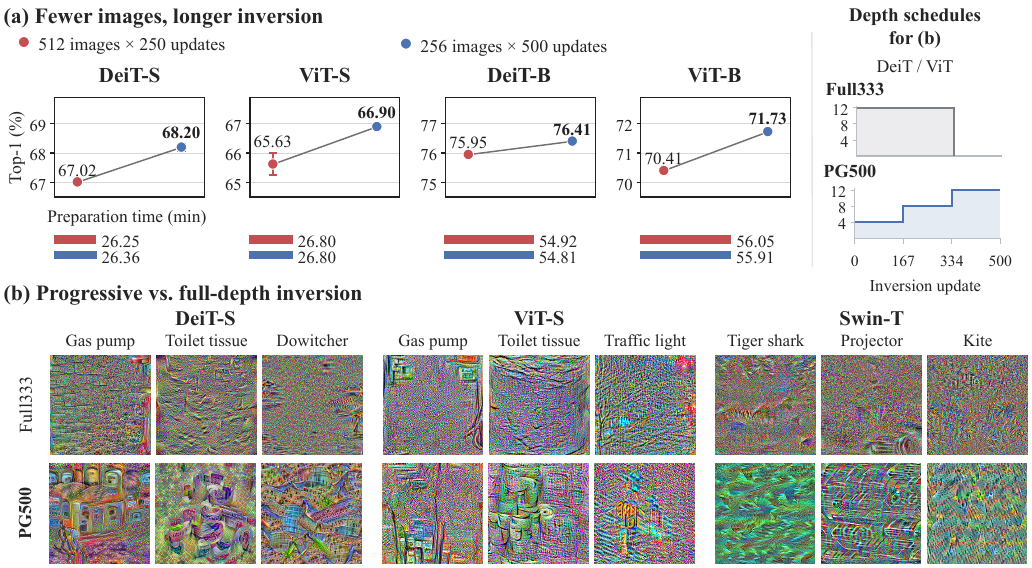}
\caption{\textbf{Allocating synthesis computation.} (a) 256 images $\times$ 500 inversion updates versus 512 $\times$ 250 at near-equal preparation cost. Points show two-seed means; error bars span the seeds. Time includes synthesis, calibration and recovery. Right: depth schedules for the DeiT/ViT examples in (b). (b) Full333/PG500 at 3,996 logical block evaluations per image, paired by initial noise and requested class. Quantitative Swin recovery uses full-depth pools.}\label{fig:synthesis-utility}
\end{figure*}

\begin{samepage}
\paragraph{Fewer images, more inversion.}
At near-equal preparation cost, 256 images synthesized for 500 updates outperform 512 images synthesized for 250 by 1.19/1.28 points on DeiT-S/ViT-S and 0.46/1.32 points on DeiT-B/ViT-B (\cref{fig:synthesis-utility}a). Longer inversion makes these smaller pools more effective.\par
\end{samepage}

\subsection{Making small synthetic pools effective}
\paragraph{Both stages contribute to recovery.}
FGC improves recovery from both Full333 and PG500 pools, while progressive synthesis (\cref{fig:synthesis-utility}b) improves recovery with either MSE or FGC preparation (\cref{tab:components}a). Their combination reaches 68.43\%/67.68\%, improving over Full333 with MSE by 3.92/1.73 points. It also exceeds equal-update Full500 by 1.01/0.72 points with 33.4\% fewer logical block evaluations.

\paragraph{Normalizing attention rows improves recovery.}
Row normalization improves recovery by 0.55/3.25 points on DeiT-S/ViT-S, with the larger gain on ViT-S (\cref{tab:components}b). The operation removes quantized row mass as an output-gain factor; recovery then adapts to this normalized representation.

The final-P-first variant installs row normalization and fixes P grids before ordinary-grid calibration, yielding 0.55/1.38-point lower accuracy than default staging (Appendix~\ref{app:staging}).

\paragraph{Better recovery from the same 256 images.}
On the 256-image PG500 pools of \cref{fig:external}, FGC yields mean gains of 1.27/0.63 points over tensor MSE on DeiT-S/ViT-S; all four paired gains are positive (\cref{tab:components}c). FGC thus improves the recovery value of an already generated pool.

\paragraph{Calibration benefits extend across backbones.}
Relative to tensor MSE, FGC improves W4A4 mean accuracy on eight of nine primary backbones and 15 of 18 paired runs; ViT-T decreases by 1.02 points. ImageNetV2 gains persist on DeiT-S, ViT-S and Swin-T (Appendix~\ref{app:v2}). 

\subsection{Why calibrate in student context?}
\paragraph{Matching the reference to the student's block state.}
Starting the reference at the student's block input gains 0.72/0.76 points over Local and 0.94/5.60 over full-teacher targets on DeiT-S/ViT-S (\cref{tab:calibration}a). These checkpoints retain 0.58/0.60-point gains over Local on ImageNetV2. A longer FP path therefore need not yield a better local grid target.

\paragraph{Interaction changes the selected grids.}
The interaction term in \cref{eq:reference-score} changes 8/12 DeiT-S and 7/12 ViT-S FC2 selections from the same candidate sets. On 64 held-out inputs, 6/8 and 7/7 changes lower contextual error, while two DeiT choices reverse (\cref{fig:reference-mechanism}).

\begin{figure}[!t]
\centering\includegraphics[width=\columnwidth]{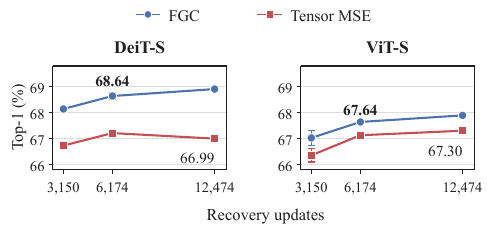}
\caption{\textbf{Prepared grids support earlier recovery.} W4A4 on the primary 1K pools along the default recovery schedule. Points show two-seed means; bars span the seeds. At 6,174 updates, FGC exceeds tensor MSE at 12,474 on both backbones.}\label{fig:recovery-checkpoints}
\end{figure}

\begin{table}[!t]\centering\small
\caption{\textbf{Scale adaptation changes the ranking of initializations.} Two-seed Top-1 means (\%) on the reference-study pools at 12,474 updates. $\Delta$ is FGC minus Local (pp). Only the ordinary-scale gradient multiplier changes.}\label{tab:reference-adapt}
\begin{tabular*}{\linewidth}{@{\extracolsep{\fill}}llrrr@{}}\toprule
Backbone & Scale & Local & FGC & $\Delta$\\\midrule
DeiT-S & Default & 67.87 & \textbf{68.59} & $+0.72$\\
 & $100\times$ & 68.78 & \textbf{69.59} & $+0.81$\\\midrule
ViT-S & Default & 67.39 & \textbf{68.15} & $+0.76$\\
 & $100\times$ & \textbf{68.82} & 68.55 & $-0.27$\\
\bottomrule\end{tabular*}

\end{table}

\paragraph{Context at nearly local-scoring cost.}
FGC calibrates DeiT-S/ViT-S in 8.32/9.06 seconds, about 0.1 seconds above Local and 1.5 seconds below block readout (\cref{tab:app-calibration-time}).

\subsection{Calibration and subsequent recovery}\label{sec:finite-recovery}
FGC reaches higher accuracy earlier in recovery (\cref{fig:recovery-checkpoints}). Along the default schedule, it achieves 68.64\%/67.64\% at 6,174 updates, exceeding tensor MSE at 12,474 updates (66.99\%/67.30\%) with 50.5\% fewer recovery updates.

With $100\times$ ordinary-scale gradients, both initializations improve, but Local rises to 68.82\% on ViT-S versus 68.55\% for FGC, reversing their default ranking. DeiT-S retains a 0.81-point contextual advantage (\cref{tab:reference-adapt}). Subsequent scale adaptation thus reshapes the value of the initial reference.

\newpage
\begin{table}[!t]\centering\small
\caption{\textbf{The low-budget advantage persists on ImageNetV2.} W4A4 Top-1 (\%). (a) Exact Fig.~1 low-budget checkpoints, two-seed means. (b) Primary 1K and external checkpoints: long-budget MimiQ uses seed 17; SARDFQ averages two complete runs, and our primary results average two recovery seeds. Amplified MimiQ includes the static-control cache.}\label{tab:external-v2}
\begin{tabular*}{\linewidth}{@{\extracolsep{\fill}}lrr@{}}\toprule
Configuration & DeiT-S & ViT-S\\\midrule
\multicolumn{3}{l}{\textit{(a) Small-budget checkpoints from Fig.~1}}\\
MimiQ 1K, grad. $100\times$ &42.72&32.81\\
\textbf{Our pipeline, 256 images} &\textbf{54.83}&\textbf{53.94}\\\midrule
\multicolumn{3}{l}{\textit{(b) Primary 1K and external checkpoints}}\\
MimiQ, native &48.49&39.96\\
MimiQ, grad. $100\times$ &54.55&50.33\\
SARDFQ, rep. off &51.27&44.27\\
\textbf{Our pipeline, primary 1K} &\textbf{55.45}&\textbf{54.95}\\\bottomrule
\end{tabular*}

\end{table}

\subsection{The small-pool advantage on ImageNetV2}
The 256-image checkpoints from \cref{fig:external} retain their advantage on ImageNetV2~\cite{imagenetv2}, reaching 54.83\%/53.94\% versus 42.72\%/32.81\% for strengthened small-budget MimiQ (\cref{tab:external-v2}a). The mean gains are 12.12/21.13 points.

The primary 1K models also exceed strengthened long-budget MimiQ by 0.90/4.62 points (\cref{tab:external-v2}b).

\section{Conclusion}
Effective data-free quantization depends on both the synthetic inputs and the student that learns from them. FGC makes block context useful for activation-grid selection through cached FP references and local candidate evaluation. Progressive inversion improves pool utility by allocating image updates across teacher depth. Together, these designs enable accurate low-bit recovery from small synthetic pools at low preparation cost, with the low-budget advantage retained on ImageNetV2.

{\small
\bibliographystyle{ieeenat_fullname}\bibliography{references}}
\clearpage
\onecolumn
\setcounter{section}{0}
\setcounter{subsection}{0}
\setcounter{figure}{0}
\setcounter{table}{0}
\setcounter{equation}{0}
\renewcommand{\theHsection}{supp.\arabic{section}}
\renewcommand{\theHfigure}{supp.\arabic{figure}}
\renewcommand{\theHtable}{supp.\arabic{table}}
\renewcommand{\theHequation}{supp.\arabic{equation}}
\section*{Supplementary Material}
\renewcommand{\thesection}{\Alph{section}}
\renewcommand{\theequation}{S\arabic{equation}}
\renewcommand{\thetable}{S\arabic{table}}
\renewcommand{\thefigure}{S\arabic{figure}}

\noindent\textbf{Guide.} \hyperref[app:protocol]{Implementation and data access}
\enspace$\cdot$\enspace\hyperref[app:budgets]{Small-pool results and timing}
\enspace$\cdot$\enspace\hyperref[app:controls]{Component and reference controls}
\enspace$\cdot$\enspace\hyperref[app:accuracy]{Full accuracy and ImageNetV2}
\enspace$\cdot$\enspace\hyperref[app:mechanism]{Grid choice and recovery}.

\section{Implementation and evaluation protocol}\label{app:protocol}
\subsection{Teachers and data access}
DeiT, ViT and Swin each contribute tiny, small and base backbones. DeiT uses ImageNet-1K checkpoints and ViT uses AugReg ImageNet-21K-to-1K checkpoints. Both have 12 blocks and 197 tokens~\cite{deit,vit}. Swin uses $7\times7$ windows, with stage depths $(2,2,6,2)$ for Swin-T and $(2,2,18,2)$ for Swin-S/B~\cite{swin}. Swin-T/S use ImageNet-1K weights and Swin-B uses ImageNet-22K-to-1K weights. Preparation uses pretrained weights, class requests and model normalization. Real images serve evaluation and analysis. ImageNet-50K informed development. On the separate ImageNetV2 MatchedFrequency split, each panel evaluates checkpoints fixed before its readout (\hyperref[app:v2]{ImageNetV2 protocol}).

\subsection{Quantized representation}
DeiT and ViT have 49 quantized linear operators, one quantized patch projection and 24 quantized attention products, with 97 activation grids. A Swin with $L$ blocks has $6L+5$ quantized operators and $8L+4$ activation grids, including three patch-merging reductions. The patch projection takes a floating-point image and quantized weights. Bias, normalization, embedding, Softmax, nonlinearities and residual operations retain their original precision. Each bitwidth starts from pretrained weights.

Let $\operatorname{clip}_{a,b}$ clip to the stated integer range. For each output channel $c$, the native symmetric weight rule uses
\begin{equation}
 r_c=\max_k|w_{ck}|,\quad s_c=\frac{\max(2r_c,10^{-8})}{2^{b_W}-1},\quad
 Q_W(w_{ck})=s_c\operatorname{clip}_{-2^{b_W-1},\,2^{b_W-1}-1}
 \left(\operatorname{round}(w_{ck}/s_c)\right).
\end{equation}
Linear channels reduce their input dimension; convolution channels flatten the input-channel and spatial dimensions. The range is recomputed from detached shadow weights in training and cached at evaluation. Weight scales are derived directly from the current shadow weights. The native custom backward passes the weight gradient straight through.

Every ordinary activation surface uses one scalar scale and offset for its whole tensor:
\begin{equation}
 Q_A(x;s,z)=s\left[\operatorname{round}\!\left(\operatorname{clip}_{q_-,q_+}(x/s+\operatorname{round}(z))\right)-\operatorname{round}(z)\right].
\end{equation}
Signed codes are $[-2^{b_A-1},2^{b_A-1}-1]$; unsigned codes are $[0,2^{b_A}-1]$. Ordinary signedness follows the native initial-view test $\min(x)<10^{-5}$ and then stays fixed. Probability grids explicitly use unsigned codes and zero offset. Tensor MSE and FGC change only legal activation scales and integer offsets.

Row normalization sums and divides in floating point. At zero row mass the fallback is uniform over legal keys, using the shifted-window mask for Swin. Rounding uses a straight-through backward pass, and clamping uses its saturation derivative.

\begin{table}[!htbp]\centering\small
\caption{Representation and optimization used by our complete preparation pipeline. Floating arithmetic is retained for fake quantization, biases, normalizations and residual operations.}\label{tab:app-representation}
\begin{tabular}{>{\raggedright\arraybackslash}p{.13\linewidth}>{\raggedright\arraybackslash}p{.24\linewidth}>{\raggedright\arraybackslash}p{.25\linewidth}>{\raggedright\arraybackslash}p{.27\linewidth}}\toprule
Surface & Granularity/support & Calibration & Recovery/inference\\\midrule
Weights & Per output channel; symmetric signed & Frozen teacher shadow weights and derived grid & Train shadow weights; dynamic min/max; cached final codes\\
Ordinary A & One affine $(s,z)$ per tensor; fixed signedness & MSE proposal then operator-score sweep & Train $s$; freeze rounded $z$\\
Probability P & One unsigned grid per surface; $z=0$ & Native-P sweep, then mean-P reset & Train scale; normalize each quantized row; zero-row fallback\\\bottomrule
\end{tabular}\end{table}

\subsection{Synthetic pools and activation-grid search}\label{app:grid}

Unless a budget control states otherwise, each default FGC pool contains $N_{\rm img}=1{,}000$ images. The flat-backbone generator requests each of the 1,000 classes once. The Swin image generator samples requests uniformly with replacement, generates 1,008 images in batches of 16, and retains the first 1,000. Full-depth synthesis (500 steps) and progressive synthesis (500 steps) use 500 AdamW image updates, betas $(0.9,0.999)$, and weight decay $10^{-6}$. The learning rate is
\begin{equation}
\eta_t=0.001+0.099[1+\cos(\pi t/T)]/2.
\end{equation}
Progressive synthesis uses $T=500$ and depths 4/8/12 for 167/167/166 updates. Full-depth synthesis (333 steps) uses $T=333$ and full depth throughout. Optimizer history continues across progressive synthesis switches. For quantitative recovery, Swin uses its full-depth image pool.

The MimiQ inversion augmentations~\cite{mimiq} apply spatial roll, channel shift/scale, and additive Gaussian noise in normalized image space. Jitter rises from eight to sixteen pixels halfway through the schedule. With the recorded modulo convention, $r_t=(T-2-t)\bmod T$ and $\sigma_t=0.5r_t/T$. Channel shifts are uniform on $[-1,1]$ and channel scales exponentiate draws from that interval. RGB projection follows each update. Classification, inter-head correlation (IHC), and total variation (TV) coefficients are 1, 1, and $2.5\times10^{-5}$. The prefix IHC term retains the full-depth denominator; deeper prefixes activate more of these terms. Starting noise and class schedule are paired within each synthesis comparison.

For calibration, 128 evenly spaced image indices supply one unaugmented view and three fixed recovery augmentations, forming a 512-view bank. Tensor sampling samples at most 2,048 values per batch and 32,768 per activation quantizer. Given scale $s_0$, proposals include the current grid, observed-range scale, and
\begin{equation}
\{s_0 10^u:u\in\operatorname{linspace}(-1.2,1.2,33)\}.
\end{equation}
For an ordinary surface, the initialization offset set is $\{z_0\}\cup(\mathbb Z\cap[q_-,q_+])$, where $z_0$ is the rounded incoming offset; P uses $\{0\}$. The incumbent is retained explicitly. Thus the additional offset proposals are legal integer codes, while the inherited grid is also evaluated. The functional sweep uses 64 evenly spaced bank indices: 16 unaugmented views and 16 from each augmented-view set. Each of these 64 scoring views uses a distinct base-image index. Given the incoming $(s,z)$ at a coordinate, refinement uses
\begin{equation}
 \mathcal C_j=\{0.875s,s,1.125s\}\times\{z,\max(q_-,z-1),\min(q_+,z+1)\},
\end{equation}
with duplicate offsets removed; P retains only $z=0$. New neighboring proposals stay within the code interval for an in-range incumbent. All tested scales are positive. A candidate replaces the current best only if its score decreases by more than $10^{-14}$, so ties retain the incumbent. Scoring converts output differences and reference responses to float64, computes their mean squared energies, and uses denominator $\operatorname{mean}(r^2)+10^{-12}$. Equivalently, summing errors and reference energies over $M$ scalar output elements gives a denominator stabilizer of $M\,10^{-12}$. Tensor collection and full-bank propagation use batches of 16; functional scoring uses all 64 views together. Units advance the full input bank after their sweep.

\paragraph{Operand refresh and sweep order.}
For each consuming-operator tap, one FP unit pass produces a target from the current student unit input. That target stays fixed until the unit boundary. Each coordinate snapshots fresh student operands by executing the current unit. Candidates share these operands and other grids. After acceptance, the implementation replays the unit and checks operand/output agreement. The bank advances through the accepted student unit.

Within a block the sweep visits QKV-projection input, attention-output-projection input, the two $QK^\top$ operands, the two $PV$ operands, MLP expansion input and MLP output-projection input, in that order. Patch merging and the final classifier have one input coordinate each. The sweep uses the native unnormalized probability operator, then resets the P scales using \cref{eq:p-init}, fixes their offsets at zero and installs row normalization. Selected ordinary grids initialize recovery.
\subsection{Recovery and pairing}\label{app:recovery}

Recovery uses batch size 16, SGD learning rate 0.001, momentum 0.9, Nesterov acceleration, and zero weight decay. For the zero-based update index $k$, the learning rate is $10^{-3}0.1^{\mathbf1[k\ge3118]+\mathbf1[k\ge6237]}$ over 12,474 updates. Teacher parameters are fixed. The optimizer includes all original student parameters: shadow weights, activation scales, biases, LayerNorm affine parameters, and positional/class-token or relative-position parameters where present. Ordinary offsets and P offsets are frozen.

Calibration and recovery use float32 model arithmetic without automatic mixed precision; float64 is used for calibration-score aggregation. Teacher and student stay in evaluation mode for dropout/stochastic-depth behavior, while the student's quantizers use their native training path during recovery. Recovery uses the native scale parameter directly, without adding a positivity projection or scale floor; positive scale proposals are a calibration constraint. Calibration performs no weight updates, and recovery begins from the saved grid state.

\paragraph{Exact recovery objective.}
For an augmented batch of $B$ images, let $z^T_i,z^S_i$ denote teacher/student logits. Set $p_i^T=\operatorname{softmax}(z_i^T/\tau)$, $p_i^S=\operatorname{softmax}(z_i^S/\tau)$. The output loss is
\begin{equation}
 \mathcal L_{\rm KD}=\frac{\tau^2}{B}\sum_{i=1}^B\sum_c p^T_{ic}\log\frac{p^T_{ic}}{p^S_{ic}},\qquad \tau=5.
\end{equation}
This is teacher-to-student KL, summed over classes and averaged over the batch, including the $\tau^2$ factor. The hard-label coefficient is zero.

HAD uses per-head $PV$ outputs before concatenation and output projection. For each layer, sample/window, head and feature channel, let $t,q$ be teacher/student token vectors, with token means $\mu_t,\mu_q$, variances $v_t=\operatorname{mean}(t^2)-\mu_t^2$, $v_q=\operatorname{mean}(q^2)-\mu_q^2$, and covariance $c_{tq}=\operatorname{mean}(tq)-\mu_t\mu_q$. The implemented structural term is
\begin{equation}
 \mathcal L_{\rm HAD}=\frac1L\sum_{l=1}^L\operatorname{mean}_{i,h,c}
 \left[1-\frac{2c_{tq}+0.03^2}{v_t+v_q+0.03^2}\right].
\end{equation}
Flat models exclude CLS and use the 196 patch tokens. Swin uses all 49 tokens per window, averages over batch-window samples, heads and channels within each layer, then averages layers equally. There is no luminance factor or extra squared-correlation term. The total loss is $\mathcal L_{\rm KD}+10\mathcal L_{\rm HAD}$.

Recovery views use RandomResizedCrop, horizontal flip, probabilistic color jitter, grayscale, and Gaussian blur. Color transforms retain original denormalization and renormalization. Teacher targets are recomputed on the actual transformed input. Recovery supervision comes from teacher responses to those views.

The native scale-gradient multiplier is $g_j=\kappa_j/\sqrt{n_jq_{+,j}}$, where $n_j$ is fixed from the initialization tensor size including batch. The saved DeiT-S/ViT-S states use $\kappa_j=1$ on 84 internal ordinary grids and $0.01$ on the classifier input. Swin-T uses 1 on 88 ordinary grids. P scales remain trainable. Seeds 17/29 vary recovery sampling and augmentation on each shared synthetic pool. ImageNet Top-1 values are fixed-checkpoint diagnostic readouts on ImageNet-50K unless stated otherwise.

\begin{table}[!htbp]\centering\small
\caption{Pool identities for DeiT-S and ViT-S at W4A4. Each FGC pair gives the two backbone means in order.}\label{tab:app-pools}
\begin{tabular}{p{.32\linewidth}p{.41\linewidth}r}\toprule
Pool & Used for & FGC Top-1\\\midrule
Primary 1K & \cref{tab:headline} and block readout & 68.911 / 67.892\\
Independent PG500, 1K & Synthesis and calibration factorial & 68.428 / 67.675\\
Matched-budget PG500, 1K & Reference origins and budget endpoints & 68.591 / 68.151\\
Actual-size PG500, 256 & Low-cost endpoint and same-pool control & 68.201 / 66.901\\
\bottomrule\end{tabular}
\end{table}

\FloatBarrier
\section{Small-pool accuracy and preparation cost}\label{app:budgets}
\subsection{Calibration on the actual 256-image pools}\label{app:256}
On each 256-image PG500 pool from \cref{fig:external}, tensor-MSE and FGC share the non-grid initial state, final probability representation, ordinary-offset policy and 12,474-update recovery schedule. The FGC endpoints are reused from the frontier. Four tensor-MSE recoveries complete the paired accuracy comparison. The frontier timing describes the original FGC preparation runs.

\begin{table}[H]\centering\small
\caption{Top-1 (\%) on the actual 256-image pools. $\Delta$ is the paired FGC gain in percentage points.}\label{tab:app-256}
\begin{tabular}{llrrr}\toprule
Model & Seed & Tensor MSE & FGC & $\Delta$ (pp)\\\midrule
\multirow{3}{*}{DeiT-S} & 17 & 66.884 & 68.082 & +1.198\\
 & 29 & 66.980 & 68.320 & +1.340\\
 & Mean & 66.932 & 68.201 & \textbf{+1.269}\\\midrule
\multirow{3}{*}{ViT-S} & 17 & 66.488 & 67.014 & +0.526\\
 & 29 & 66.058 & 66.788 & +0.730\\
 & Mean & 66.273 & 66.901 & \textbf{+0.628}\\\bottomrule
\end{tabular}

\end{table}

\subsection{Image counts and inversion duration}\label{app:budget-matrix}
Each 256- and 512-image pool is generated at its actual size. The 250-update schedule uses 84/83/83 updates at depths 4/8/12 (1,996 logical block evaluations per image), with the learning rate and augmentations retimed to $T=250$. The 500-update schedule uses 3,996 logical block evaluations per image. Recovery uses 12,474 updates in both cases. Within each backbone the budget matrix uses the same teacher and recovery recipe as the architecture comparison, with its own pool realization. At near-equal preparation cost, 256/500 improves upon 512/250 on all four backbones in \cref{fig:synthesis-utility}. DeiT-B peaks at 512/500 in this grid. ViT-B peaks at 1K/500.

\begin{table}[!htbp]\centering\small
\caption{All four backbone budget grids. Each cell gives two-seed mean Top-1 (\%) and complete preparation time (min).}\label{tab:app-budget}
\begin{tabular}{rrrrrr}\toprule
Images & Updates & DeiT-S & ViT-S & DeiT-B & ViT-B\\\midrule
256 & 250 & 67.003 / 18.84 & 65.228 / 19.31 & 75.909 / 38.95 & 69.190 / 39.67\\
256 & 500 & 68.201 / 26.36 & 66.901 / 26.80 & 76.407 / 54.81 & 71.730 / 55.91\\
512 & 250 & 67.016 / 26.25 & 65.626 / 26.80 & 75.952 / 54.92 & 70.407 / 56.05\\
512 & 500 & 68.411 / 41.20 & 66.887 / 41.78 & 76.691 / 86.62 & 70.250 / 88.07\\
1000 & 250 & 67.275 / 40.53 & 65.943 / 41.00 & 75.873 / 84.94 & 67.308 / 86.01\\
1000 & 500 & 68.591 / 69.34 & 68.151 / 70.44 & 76.356 / 146.65 & 73.417 / 148.21\\
\bottomrule\end{tabular}
\end{table}

\subsection{External protocols and stage costs}\label{app:external}
We measure wall time on the same RTX 5090 for each comparison, synchronizing at stage boundaries and excluding process startup and final evaluation. Shared synthesis is charged in full to every recovered model. FGC logs calibration separately. The MimiQ recovery timer includes its native initialization and calibration, and the SARDFQ reconstruction timer includes calibration.

\paragraph{Small-budget MimiQ.}\label{app:mimiq-small}
Native batch-16 synthesis generates 1,008 images with 500 full-depth updates and retains 1,000. A complete 200-epoch schedule includes two initialization epochs and $198\times63=12{,}474$ SGD updates. The native and amplified runs within each model share the pool and initialization seeds, use native min/max weights, first-batch activation initialization, native P and offset behavior, temperature-5 KD and HAD10. Both use the fixed-control cache. The amplified version applies $100\times$ only to ordinary-scale gradients and supplies the two-seed means in \cref{fig:external}. Both variants use the same one-thread execution policy. The preparation timer charges all 1,008 generated images.

\paragraph{Long-budget MimiQ.}\label{app:mimiq-long}
The released-code and strengthened configurations use 10K images, 2,000 full-depth inversion updates and 123,750 student updates (seed 17). They share a batch size of 16, a 200-epoch recovery schedule and the same measured synthesis stage. The strengthened configuration combines $100\times$ ordinary-scale gradients with the cache below. Published W4A4 LSQ accuracy is 62.72\%/55.69\% for DeiT-S/ViT-S, compared with 62.314\%/51.492\% in our native reproduction. \Cref{tab:headline} uses reported results, and \cref{fig:external} pairs each reproduced checkpoint with its own measured time. The released code initializes activations from the first batch using the absolute-mean LSQ rule, while the paper describes min/max initialization. The local runs use HAD weight 10.

\paragraph{SARDFQ.}
The DeiT-S/ViT-S points use 32 images, 1,000 generation updates, model-specific attention priors, native semantic reinforcement and log2 P quantization. With reparameterization off, recovery uses 100 updates per unit per reconstruction stage and two complete runs per model. The separate Swin-T adaptation in \cref{tab:headline} uses the code-default 1,000 reconstruction updates per unit and stage, the original reparameterization switch and the Swin-S attention-prior coefficient 10. Its W4A4 seed pair is 69.708/67.764 and W5A5 pair is 75.542/76.082.

\begin{table}[!htbp]\centering\small
\caption{Measured preparation stages for representative Fig.~1 points (min). FGC rows use 500 inversion updates. For MimiQ and SARDFQ, recovery includes calibration. Long-budget MimiQ recovery is the reported total less the shared measured synthesis time. Dashes denote an unavailable two-run stage split. Top-1 is diagnostic ImageNet-50K accuracy (\%).}\label{tab:app-stage-costs}
\begin{tabular}{llrrrrrr}\toprule
Backbone & Method & Runs & Synthesis & Cal. & Recovery & Total & Top-1\\\midrule
DeiT-S & FGC, 256 / 500 & 2 & 14.99 & 0.12 & 11.24 & 26.36 & 68.201\\
DeiT-S & FGC, 1000 / 500 & 2 & -- & -- & -- & 69.34 & 68.591\\
DeiT-S & MimiQ, small $100\times$ & 2 & 17.87 & incl. & 10.34 & 28.20 & 55.173\\
DeiT-S & MimiQ, long native & 1 & 714.37 & incl. & 115.64 & 830.01 & 62.314\\
DeiT-S & MimiQ, long $100\times$ + cache & 1 & 714.37 & incl. & 104.49 & 818.85 & 67.572\\
DeiT-S & SARDFQ & 2 & 2.35 & incl. & 1.46 & 3.80 & 64.028\\
\midrule
ViT-S & FGC, 256 / 500 & 2 & 15.16 & 0.13 & 11.51 & 26.80 & 66.901\\
ViT-S & FGC, 1000 / 500 & 2 & -- & -- & -- & 70.44 & 68.151\\
ViT-S & MimiQ, small $100\times$ & 2 & 20.34 & incl. & 23.61 & 43.95 & 42.537\\
ViT-S & MimiQ, long native & 1 & 723.39 & incl. & 119.27 & 842.66 & 51.492\\
ViT-S & MimiQ, long $100\times$ + cache & 1 & 723.39 & incl. & 105.43 & 828.82 & 62.396\\
ViT-S & SARDFQ & 2 & 2.36 & incl. & 1.50 & 3.86 & 56.572\\
\bottomrule\end{tabular}
\end{table}

\paragraph{Static quantizer controls.}\label{app:cache}
After native initialization, the cache specializes the fixed flags \texttt{init\_flag} and \texttt{signed}. Learned scales, offsets and dynamic weight ranges follow the original arithmetic. A 256-update continuation from the same amplified checkpoint and 10K pool used matched reset optimizers and deterministic random states. Augmented inputs, forward outputs, first-step gradients and final model tensors matched across native and cached arms. With four warmup updates and native runs bracketing the cached arm, the cached arm reduces seconds per update by 10.47\% on DeiT-S and 9.87\% on ViT-S.

\begin{table}[!htbp]\centering\small
\caption{MimiQ control-cache probe. Each cell reports seconds per update.}\label{tab:app-cache}
\begin{tabular}{lrrr}\toprule
Model & Native 1 & Cached & Native 2\\\midrule
DeiT-S & 0.05724 & 0.05144 & 0.05766\\
ViT-S & 0.06007 & 0.05377 & 0.05925\\
\bottomrule\end{tabular}
\end{table}

\FloatBarrier
\section{Synthesis, reference and probability controls}\label{app:controls}
\subsection{Synthesis and calibration factorial}
Full333 and PG500 each use 3,996 logical block evaluations per image. The four cells cross the two independent 1K pools with tensor-MSE and FGC. Both initializers share the final P operator, frozen offsets and recovery schedule. Full500 adds the equal-update reference. At a fixed saved Full333 FGC initializer, switching only recovery images to PG500 raises DeiT-S/ViT-S to 68.200\%/67.456\%. Switching the calibration pool as well reaches 68.428\%/67.675\%. Those increments describe this ordered crossover.

\begin{table}[!htbp]\centering\small
\caption{Four matched-work synthesis/calibration cells. W4A4 ImageNet Top-1 (\%) over seeds 17 and 29.}\label{tab:app-factorial}
\begin{tabular}{llrrrrrr}\toprule
&&\multicolumn{3}{c}{DeiT-S}&\multicolumn{3}{c}{ViT-S}\\
Synthesis & Preparation & Seed17 & Seed29 & Mean & Seed17 & Seed29 & Mean\\\midrule
Full333 & Tensor MSE & 64.416 & 64.610 & 64.513 & 66.058 & 65.834 & 65.946\\
Full333 & FGC & 66.510 & 66.404 & 66.457 & 66.540 & 66.616 & 66.578\\
PG500 & Tensor MSE & 66.958 & 67.082 & 67.020 & 67.090 & 66.812 & 66.951\\
PG500 & FGC & 68.498 & 68.358 & 68.428 & 67.810 & 67.540 & 67.675\\
\bottomrule\end{tabular}

\end{table}

\begin{table}[!htbp]\centering\small
\caption{Fixed-initializer crossover and equal-update reference. W4A4 ImageNet Top-1 (\%).}\label{tab:app-crossover}
\begin{tabular}{llrrr}\toprule
Control & Model & Seed 17 & Seed 29 & Mean\\\midrule
Full-depth calibration / progressive recovery & DeiT-S & 68.154 & 68.246 & 68.200\\
Full-depth calibration / progressive recovery & ViT-S & 67.752 & 67.160 & 67.456\\
Full-depth synthesis (500 steps) / FGC & DeiT-S & 67.504 & 67.340 & 67.422\\
Full-depth synthesis (500 steps) / FGC & ViT-S & 67.124 & 66.794 & 66.959\\
\bottomrule\end{tabular}
\end{table}

\begin{figure}[!htbp]\centering
\includegraphics[width=\linewidth]{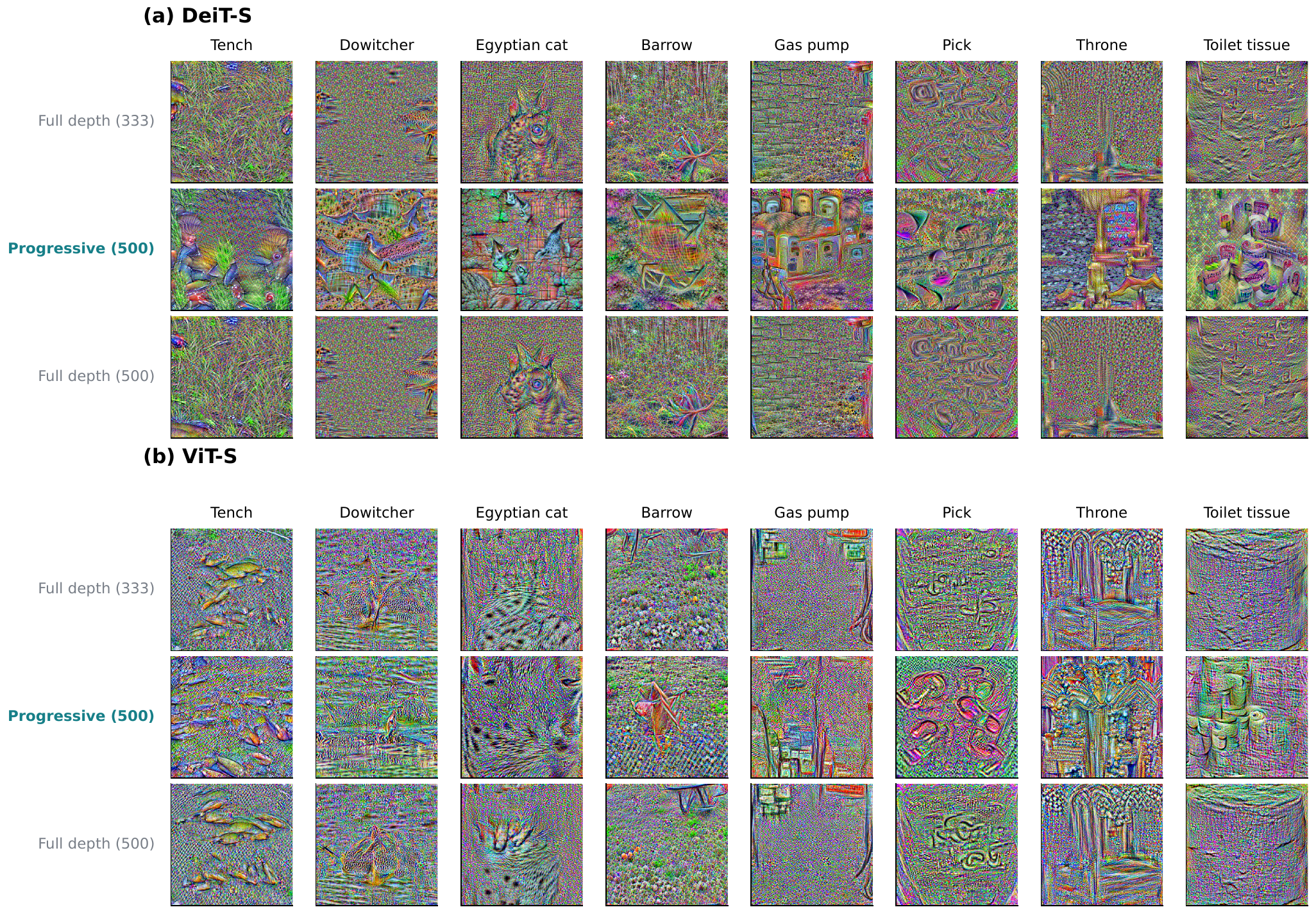}
\caption{Paired synthetic endpoints for fixed class IDs. Full333 and PG500 match logical block work, and Full500 and PG500 match the number of image updates.}\label{fig:app-synthesis}
\end{figure}

The paired examples in \cref{fig:app-synthesis} use eight fixed class IDs from the same independent DeiT-S and ViT-S cohorts. They add the equal-update Full500 reference to the Full333/PG500 comparison. Each column keeps the teacher, class and initial Gaussian sample fixed. Images use the teacher's original denormalization.

For Swin-T the examples in \cref{fig:synthesis-utility} share the noise and class requests across schedules. The progressive schedule uses 4/8/12 blocks for 167/167/166 updates and retains the full-depth IHC denominator. At truncated depths, native merging feeds the final normalization, pooling and classifier. The count of 3,996 is logical block work. Quantitative Swin recovery uses full-depth pools.

\FloatBarrier
\subsection{Reference origin and readout}\label{app:reference}
The primary-pool and matched-budget comparisons replace the reference while keeping the candidate rule and consuming-operator readout fixed. Local applies the FP consuming operator to current student operands. The full-teacher variant caches taps along the FP trajectory from the image. FGC anchors the FP unit at the current student unit input. Candidate operands are fixed within a coordinate and refreshed after earlier commits. Local and FGC here share the tensor proposals, final P operator, frozen offsets and subsequent recovery schedule.

\begin{table}[!htbp]\centering\small
\caption{Per-seed reference-origin controls. The primary and budget studies use distinct 1K pools. Top-1 (\%).}\label{tab:app-reference}
\begin{tabular}{lllrrr}\toprule
Model & Pool & FP target & Seed 17 & Seed 29 & Mean\\\midrule
DeiT-S & Budget 1K & Student block input (default) & 68.564 & 68.618 & 68.591\\
DeiT-S & Budget 1K & Local operator reference & 67.932 & 67.804 & 67.868\\
DeiT-S & Budget 1K & Full FP trajectory & 67.462 & 67.834 & 67.648\\
ViT-S & Budget 1K & Student block input (default) & 68.328 & 67.974 & 68.151\\
ViT-S & Budget 1K & Local operator reference & 67.364 & 67.420 & 67.392\\
ViT-S & Budget 1K & Full FP trajectory & 62.602 & 62.506 & 62.554\\
\midrule
DeiT-S & Primary 1K & Local & 68.232 & 68.150 & 68.191\\
DeiT-S & Primary 1K & Student block & 68.802 & 69.020 & 68.911\\
ViT-S & Primary 1K & Local & 67.140 & 67.250 & 67.195\\
ViT-S & Primary 1K & Student block & 67.782 & 68.002 & 67.892\\
\bottomrule\end{tabular}
\end{table}

Block readout compares each candidate's complete student-unit output with the FP unit output at the same student unit input. It normalizes the squared error by the FP output energy over the scoring bank with the same mean-energy stabilizer as the operator score. The full student unit runs for each candidate. In \cref{tab:calibration}b the FP reference origin, proposals and recovery are common to both readouts.

\begin{table}[!htbp]\centering\small
\caption{Primary W4A4 readout ablation with the FP student-input reference. ImageNet Top-1 (\%).}\label{tab:app-readout}
\begin{tabular}{llrrr}\toprule
Model & Readout & Seed 17 & Seed 29 & Mean\\\midrule
DeiT-S & Operator & 68.802 & 69.020 & 68.911\\
DeiT-S & Block & 68.588 & 68.648 & 68.618\\
Swin-T & Operator & 75.110 & 74.848 & 74.979\\
Swin-T & Block & 75.098 & 74.538 & 74.818\\
ViT-S & Operator & 67.782 & 68.002 & 67.892\\
ViT-S & Block & 67.650 & 68.000 & 67.825\\
\bottomrule\end{tabular}
\end{table}

\subsection{Calibration time}\label{app:calibration-time}
For each objective and model we interleave three complete calibration-stage repetitions on one GPU. The timer starts after imports and covers model preparation, proposals, refinement and checkpoint/replay. Views and candidates match across objectives.

\begin{table}[!htbp]\centering\small
\caption{Calibration-stage seconds, mean $\pm$ sample standard deviation over three repetitions.}\label{tab:app-calibration-time}
\begin{tabular}{llr}\toprule
Model & Objective & Time (s)\\\midrule
DeiT-S & Local-reference operator & $8.216\pm0.078$\\
DeiT-S & Contextual operator (FGC) & $8.324\pm0.049$\\
DeiT-S & Contextual block readout & $9.820\pm0.073$\\
ViT-S & Local-reference operator & $8.961\pm0.057$\\
ViT-S & Contextual operator (FGC) & $9.060\pm0.069$\\
ViT-S & Contextual block readout & $10.604\pm0.025$\\
\bottomrule\end{tabular}
\end{table}

\FloatBarrier
\subsection{Probability normalization and installation order}\label{app:staging}
The normalization control branches from the same MSE initializer on the matched 1K pool, holds the mean-P scale and recovery schedule fixed, and toggles row normalization during recovery and inference. For a nonzero row $q$ with mass $m$, the normalized output $(q/m)\widehat V$ removes the multiplicative mass $m$ from the value aggregation.

Final-P-first installs the final P scales and normalization before the functional sweep, freezes the 12 P coordinates and searches the 85 ordinary grids from the same scalar proposals. The late-installation control skips P search and retains native P while scoring ordinary grids. Both finish with the same recovery operator. The final-P-first means are 0.546/1.377 points lower than default staging on DeiT-S/ViT-S.

\begin{table}[!htbp]\centering\small
\caption{Row normalization at a shared MSE initializer. ImageNet Top-1 (\%).}\label{tab:app-norm}
\begin{tabular}{lrrrrrr}\toprule
Model & OFF 17 & OFF 29 & ON 17 & ON 29 & OFF mean & ON mean\\\midrule
DeiT-S & 66.234 & 66.294 & 66.836 & 66.792 & 66.264 & 66.814\\
ViT-S & 63.944 & 63.454 & 67.206 & 66.688 & 63.699 & 66.947\\
\bottomrule\end{tabular}

\end{table}

\begin{table}[!htbp]\centering\small
\caption{Probability stages on the matched-budget pools. ImageNet Top-1 (\%).}\label{tab:app-staging}
\begin{tabular}{llrrr}\toprule
Model & P staging & Seed 17 & Seed 29 & Mean\\\midrule
DeiT-S & Student block input (default) & 68.564 & 68.618 & 68.591\\
DeiT-S & Skip P search; install final P late & 68.566 & 68.534 & 68.550\\
DeiT-S & Final-P-first ordinary sweep & 67.970 & 68.120 & 68.045\\
ViT-S & Student block input (default) & 68.328 & 67.974 & 68.151\\
ViT-S & Skip P search; install final P late & 67.922 & 67.786 & 67.854\\
ViT-S & Final-P-first ordinary sweep & 66.846 & 66.702 & 66.774\\
\bottomrule\end{tabular}
\end{table}

\FloatBarrier
\section{Architecture results and ImageNetV2}\label{app:accuracy}
\subsection{ImageNet-50K across architectures}
\Cref{tab:app-headline} extends the main comparison to the general-purpose methods under their published protocols. \Cref{tab:app-nine} gives the matched W4A4 calibration comparison on the nine primary pools. FGC improves eight backbone means and 15 of 18 seed pairs. ViT-T decreases by 1.022 points.

\begin{table}[!htbp]\centering\footnotesize
\caption{Complete historical ImageNet Top-1 comparison (\%). Source and missing-cell conventions follow \cref{tab:headline}.}\label{tab:app-headline}
\begin{tabular*}{\textwidth}{@{\extracolsep{\fill}}llrrrrrrrrr@{}}\toprule
 & & \multicolumn{3}{c}{ViT} & \multicolumn{3}{c}{DeiT} & \multicolumn{3}{c}{Swin}\\
\cmidrule(lr){3-5}\cmidrule(lr){6-8}\cmidrule(lr){9-11}
W/A & Method & T & S & B & T & S & B & T & S & B\\\midrule
FP & Local teacher & 75.34 & 81.38 & 84.41 & 72.02 & 79.72 & 81.74 & 80.92 & 82.90 & 84.75\\\midrule
4/4 & GDFQ$^{\ddagger}$ & 2.95 & 4.62 & 11.73 & 25.96 & 22.12 & 30.04 & 42.08 & 41.93 & 36.04\\
 & Qimera$^{\ddagger}$ & 0.57 & 7.02 & 5.61 & 15.18 & 11.37 & 32.49 & 47.98 & 39.64 & 29.27\\
 & AdaDFQ$^{\ddagger}$ & 2.00 & 1.78 & 6.21 & 19.57 & 14.44 & 19.22 & 38.88 & 39.40 & 32.26\\
\addlinespace[2pt]
 & PSAQ-ViT$^{\ddagger}$ & 0.67 & 0.15 & 0.94 & 19.61 & 5.90 & 8.74 & 22.71 & 9.26 & 23.69\\
 & PSAQ-ViT V2$^{\ddagger}$ & 1.54 & 4.14 & 2.83 & 22.82 & 32.57 & 45.81 & 50.42 & 39.10 & 39.26\\
\addlinespace[2pt]
 & MimiQ & 42.99 & \underline{55.69} & 62.91 & 52.03 & 62.72 & 74.10 & 69.33 & 70.46 & 73.49\\
 & SARDFQ & -- & 50.32 & 51.84 & 52.06 & 62.29 & 72.17 & 68.74$^{\dagger}$ & \underline{74.74} & \underline{76.42}\\
 & MaskAQ & \textbf{44.90} & -- & \underline{70.53} & \underline{53.10} & 63.81 & \underline{74.41} & \underline{70.40} & -- & --\\
 & CLAMP-ViT & -- & -- & -- & -- & \textbf{69.01} & -- & -- & -- & --\\
 & Ours & \underline{44.66} & \textbf{67.89} & \textbf{71.26} & \textbf{58.62} & \underline{68.91} & \textbf{76.48} & \textbf{74.98} & \textbf{78.67} & \textbf{79.59}\\
\midrule
5/5 & GDFQ$^{\ddagger}$ & 24.40 & 53.96 & 33.56 & 44.76 & 57.00 & 71.03 & 61.30 & 78.04 & 70.55\\
 & Qimera$^{\ddagger}$ & 26.70 & 16.13 & 9.43 & 33.13 & 33.65 & 47.01 & 62.13 & 46.81 & 43.57\\
 & AdaDFQ$^{\ddagger}$ & 27.10 & 59.36 & 43.02 & 53.85 & 59.55 & 71.12 & 64.61 & 79.82 & 75.59\\
\addlinespace[2pt]
 & PSAQ-ViT$^{\ddagger}$ & 17.66 & 23.37 & 16.80 & 53.36 & 47.35 & 57.23 & 58.63 & 76.33 & 57.80\\
 & PSAQ-ViT V2$^{\ddagger}$ & 40.21 & 63.59 & 74.29 & 55.18 & 65.30 & 73.16 & 69.77 & 80.55 & 79.80\\
\addlinespace[2pt]
 & MimiQ & 62.40 & 70.02 & 78.09 & 63.40 & 72.59 & 78.20 & 76.39 & \underline{80.75} & \underline{82.05}\\
 & SARDFQ & -- & \underline{74.31} & 70.70 & \underline{66.41} & \underline{74.06} & 78.72 & 75.81$^{\dagger}$ & 79.56 & 80.82\\
 & MaskAQ & \underline{63.01} & -- & \underline{78.60} & 64.27 & 73.18 & \underline{78.89} & \underline{76.82} & -- & --\\
 & Ours & \textbf{65.91} & \textbf{76.03} & \textbf{80.50} & \textbf{67.50} & \textbf{76.45} & \textbf{79.90} & \textbf{78.73} & \textbf{81.29} & \textbf{82.13}\\
\bottomrule\end{tabular*}

\end{table}

\begin{table}[!htbp]\centering\small
\caption{Matched W4A4 ImageNet Top-1 (\%) on primary pools. Each pair lists seeds 17/29. $\Delta$ is FGC minus tensor MSE (pp).}\label{tab:app-nine}
\begin{tabular}{lrr|rr|r}\toprule
Model & MSE pair & Mean & FGC pair & Mean & $\Delta$\\\midrule
ViT-T & 45.716/45.648 & 45.682 & 44.642/44.678 & 44.660 & -1.022\\
ViT-S & 67.370/67.228 & 67.299 & 67.782/68.002 & 67.892 & +0.593\\
ViT-B & 69.142/68.894 & 69.018 & 71.300/71.224 & 71.262 & +2.244\\
DeiT-T & 57.910/57.778 & 57.844 & 58.512/58.730 & 58.621 & +0.777\\
DeiT-S & 66.950/67.034 & 66.992 & 68.802/69.020 & 68.911 & +1.919\\
DeiT-B & 76.062/76.094 & 76.078 & 76.496/76.458 & 76.477 & +0.399\\
Swin-T & 74.318/74.316 & 74.317 & 75.110/74.848 & 74.979 & +0.662\\
Swin-S & 78.752/78.342 & 78.547 & 78.732/78.616 & 78.674 & +0.127\\
Swin-B & 79.612/79.450 & 79.531 & 79.702/79.470 & 79.586 & +0.055\\
\bottomrule\end{tabular}
\end{table}

\begin{table}[!htbp]\centering\small
\caption{W5A5 FGC on primary pools. Top-1 (\%) at the final checkpoint.}\label{tab:app-w5}
\begin{tabular}{lrrr}\toprule
Model & Seed 17 & Seed 29 & Mean\\\midrule
DeiT-B & 79.896 & 79.910 & 79.903\\
DeiT-S & 76.382 & 76.522 & 76.452\\
DeiT-T & 67.554 & 67.446 & 67.500\\
Swin-B & 82.140 & 82.118 & 82.129\\
Swin-S & 81.304 & 81.276 & 81.290\\
Swin-T & 78.892 & 78.568 & 78.730\\
ViT-B & 80.478 & 80.514 & 80.496\\
ViT-S & 76.042 & 76.026 & 76.034\\
ViT-T & 65.874 & 65.950 & 65.912\\
\bottomrule\end{tabular}
\end{table}

\paragraph{Additional precisions.}
The same operator-readout recipe yields widely different W3A3 accuracy across the three reference backbones. Increasing activation precision at fixed W4 also improves the reported means in \cref{tab:app-precision}.

\begin{table}[!htbp]\centering\small
\caption{Fixed-recipe bitwidth sweep. ImageNet Top-1 (\%), means over two seeds. P equals A.}\label{tab:app-precision}
\begin{tabular}{lrrrrrrr}\toprule
Model & W3A3 & W4A4 & W5A5 & W6A6 & W7A7 & W4A8 & W8A8\\\midrule
DeiT-S & 31.678 & 68.911 & 76.452 & 78.217 & 78.978 & 77.963 & 79.340\\
ViT-S & 15.367 & 67.892 & 76.034 & 79.318 & 80.413 & 77.935 & 80.666\\
Swin-T & 54.161 & 74.979 & 78.730 & 80.087 & 80.526 & 78.838 & 80.627\\
\bottomrule\end{tabular}

\end{table}

\FloatBarrier
\subsection{Frozen-checkpoint evaluation on ImageNetV2}\label{app:v2}
We evaluate frozen checkpoints on all 10,000 MatchedFrequency inputs~\cite{imagenetv2}, with PIL bilinear Resize256/CenterCrop224, model normalization and the original top-five tie rule. Each panel fixes its checkpoints before its readout. Before the first query, the evaluation fixed three FP teachers and 18 primary student checkpoints from MSE, block readout and FGC. The first panel reports the student results. The reference-origin panel uses the separate budget 1K pools with default recovery. The long-budget MimiQ and SARDFQ rows use the external endpoints in \cref{fig:external}. The 256-image comparison uses the exact FGC and strengthened small-budget MimiQ frontier checkpoints with 12,474 recovery updates.

\begin{table}[!htbp]\centering\small
\caption{Primary pools. W4A4 ImageNetV2 Top-1 (\%) at the same checkpoints used for the primary comparisons.}\label{tab:app-v2-primary}
\begin{tabular}{llrrr}\toprule
Model & Preparation & Seed 17 & Seed 29 & Mean\\\midrule
DeiT-S & Tensor MSE & 53.930 & 53.690 & 53.810\\
DeiT-S & FGC: block readout & 55.700 & 55.330 & 55.515\\
DeiT-S & FGC & 55.520 & 55.370 & 55.445\\
ViT-S & Tensor MSE & 54.360 & 54.190 & 54.275\\
ViT-S & FGC: block readout & 55.140 & 54.670 & 54.905\\
ViT-S & FGC & 54.830 & 55.060 & 54.945\\
Swin-T & Tensor MSE & 61.020 & 61.030 & 61.025\\
Swin-T & FGC: block readout & 62.030 & 61.390 & 61.710\\
Swin-T & FGC & 61.500 & 61.320 & 61.410\\
\bottomrule\end{tabular}
\end{table}

\begin{table}[!htbp]\centering\small
\caption{Matched-budget 1K pools. ImageNetV2 Top-1 (\%) for the reference-origin study.}\label{tab:app-v2-reference}
\begin{tabular}{llrrr}\toprule
Model & Preparation / reference & Seed 17 & Seed 29 & Mean\\\midrule
DeiT-S & Tensor MSE & 53.790 & 53.850 & 53.820\\
DeiT-S & Local & 54.900 & 54.950 & 54.925\\
DeiT-S & Full teacher & 54.240 & 55.210 & 54.725\\
DeiT-S & FGC & 55.620 & 55.390 & 55.505\\
ViT-S & Tensor MSE & 54.410 & 54.140 & 54.275\\
ViT-S & Local & 54.530 & 54.190 & 54.360\\
ViT-S & Full teacher & 50.420 & 49.790 & 50.105\\
ViT-S & FGC & 55.170 & 54.740 & 54.955\\
\bottomrule\end{tabular}

\end{table}

\begin{table}[!htbp]\centering\small
\caption{External long-budget MimiQ and SARDFQ checkpoints from the accuracy--cost comparison. The primary FGC comparator is in \cref{tab:app-v2-primary}. ImageNetV2 Top-1 (\%).}\label{tab:app-v2-external}
\begin{tabular}{llrr}\toprule
Configuration & Seed & DeiT-S & ViT-S\\\midrule
MimiQ, native & 17 & 48.490 & 39.960\\
MimiQ, scale grad. $100\times$ & 17 & 54.550 & 50.330\\
SARDFQ, rep. off & 17 & 50.700 & 43.570\\
SARDFQ, rep. off & 29 & 51.840 & 44.960\\
\bottomrule\end{tabular}
\end{table}

\begin{table}[!htbp]\centering\small
\caption{Exact low-cost frontier checkpoints on ImageNetV2. Top-1 (\%).}\label{tab:app-v2-small}
\begin{tabular}{llrrr}\toprule
Model & Configuration & Seed17 & Seed29 & Mean\\\midrule
DeiT-S & MimiQ 1K/Full500, grad. $100\times$ & 42.740 & 42.690 & 42.715\\
DeiT-S & Our pipeline 256/PG500 & 54.880 & 54.780 & 54.830\\
ViT-S & MimiQ 1K/Full500, grad. $100\times$ & 33.010 & 32.610 & 32.810\\
ViT-S & Our pipeline 256/PG500 & 54.340 & 53.530 & 53.935\\
\bottomrule\end{tabular}

\end{table}

\FloatBarrier
\section{Why grids change and how recovery adapts}\label{app:mechanism}
\subsection{Candidate ranking at a fixed state}
For a candidate $\eta$, let $e_\eta$ denote its difference from the local FP response, and let $b$ denote the local FP response minus the contextual FP response, as in \cref{eq:reference-score}. Within a coordinate search the operands upstream of that coordinate are fixed, so $b$ is shared across candidates. If candidate 2 has greater squared local error, the contextual score prefers it when
\begin{equation}
 2\langle e_2-e_1,b\rangle<-(\|e_2\|^2-\|e_1\|^2).
\end{equation}
When the two FP responses coincide, $b=0$ and the rankings coincide. For a finite example, take $u_T=1$ and $u_Q=1.4$ from an unsigned two-bit grid of scale 1.4. For an identity operator, candidate scales 1 and 0.5 give outputs 1 and 1.5. The local squared errors are 0.16 and 0.01. The contextual errors are 0 and 0.25. The two scores select different candidates.

\begin{figure}[!htbp]\centering
\textbf{DeiT-S}\\[2pt]
\includegraphics[width=.94\linewidth]{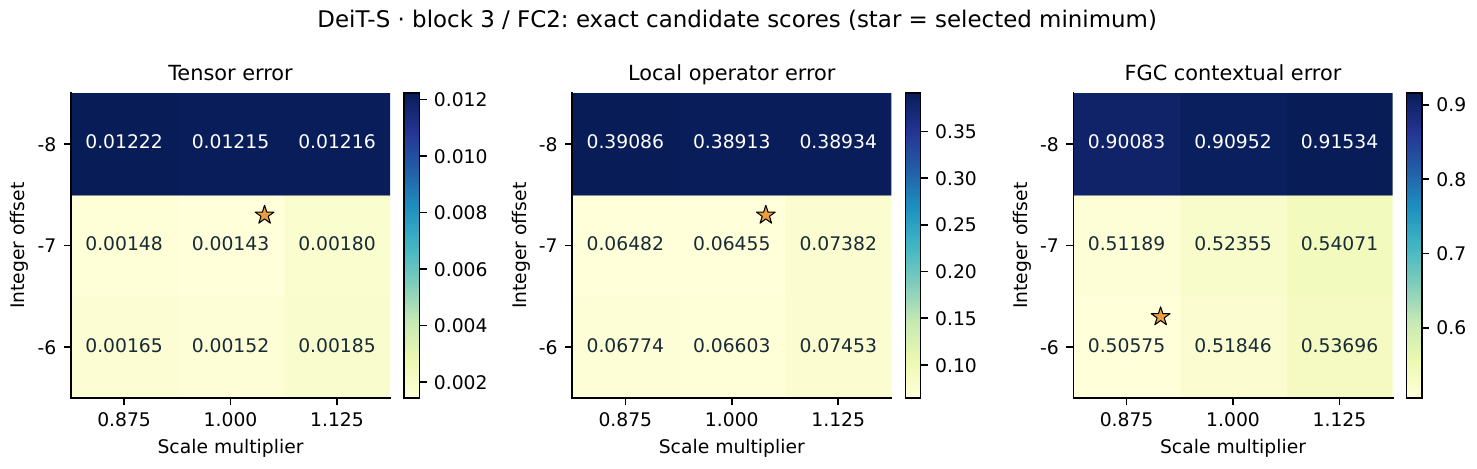}\\[4pt]
\textbf{ViT-S}\\[2pt]
\includegraphics[width=.94\linewidth]{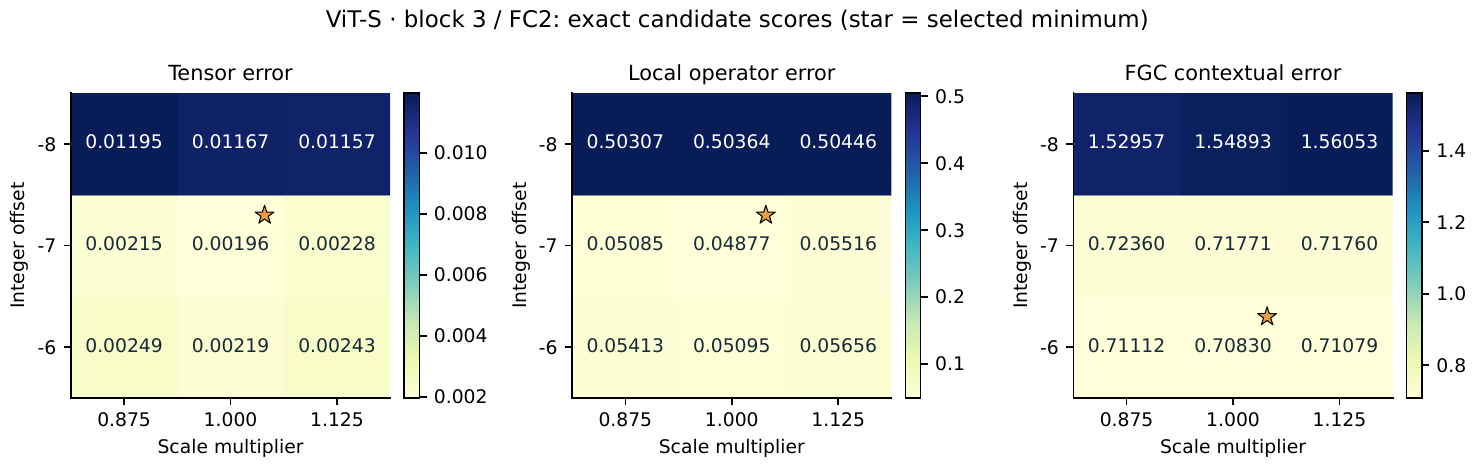}
\caption{The same legal FC2 candidate grids under three objectives for each backbone. Stars mark each score minimum. Color scales are specific to each objective.}\label{fig:app-scores}
\end{figure}

In \cref{fig:app-scores}, each objective scores the same candidates at block 3 FC2. The tensor score is computed at that coordinate state. The pipeline's initializer computes its tensor proposals earlier. Across all twelve FC2 coordinates, Local and FGC select on the same 64 views using their respective scores. On 64 calibration-excluded synthetic images, the new choice reduces contextual error in 6 of 8 changed DeiT-S coordinates and all 7 changed ViT-S coordinates. Two DeiT-S choices increase the error. This replay uses native P during calibration, before the post-sweep reset.
On the frozen tensor-MSE and FGC initializers, a separate end-to-end diagnostic holds weights and the final P operator fixed. Mean final-block patch error decreases on DeiT-S and increases on ViT-S across the four preset synthetic and real-view groups. ViT-S still improves in recovered classification. The operator score evaluates one candidate at the current state. Recovery then updates weights and activation scales.

\FloatBarrier
\subsection{Scale adaptation and finite-budget recovery}
The $100\times$ intervention multiplies the actual ordinary-scale gradient after backward, with the final probability operator, saved initializers and recovery budget fixed. For the flat-model classifier scale, it changes the effective coefficient from 0.01 to 1. Stronger adaptation reduces the FGC--MSE gap most on ViT-S and preserves the largest gain on DeiT-S. FGC still leads Local on DeiT-S, while Local overtakes FGC on ViT-S in both recovery seeds. The flat-backbone controls use the matched-budget 1K pools, while Swin-T uses its full-depth pool. The amplified Local and contextual arms include a verified cross-host accuracy comparison.

\begin{table}[!htbp]\centering\small
\caption{MSE and FGC under native and amplified ordinary-scale gradients. ImageNet Top-1 (\%) with seed pairs 17/29. Gap is FGC minus MSE (pp).}\label{tab:app-scale}
\begin{tabular}{llrrrrr}\toprule
Model & Ordinary-scale rule & MSE pair & FGC pair & MSE mean & FGC mean & Gap\\\midrule
DeiT-S & Native & 66.836/66.792 & 68.564/68.618 & 66.814 & 68.591 & +1.777\\
 & $100\times$ & 68.140/68.176 & 69.616/69.572 & 68.158 & 69.594 & +1.436\\
ViT-S & Native & 67.206/66.688 & 68.328/67.974 & 66.947 & 68.151 & +1.204\\
 & $100\times$ & 68.494/68.482 & 68.522/68.570 & 68.488 & 68.546 & +0.058\\
Swin-T & Native & 74.318/74.316 & 75.110/74.848 & 74.317 & 74.979 & +0.662\\
 & $100\times$ & 74.552/74.842 & 74.954/74.962 & 74.697 & 74.958 & +0.261\\\bottomrule
\end{tabular}
\end{table}

\begin{table}[!htbp]\centering\small
\caption{Local and FGC under each scale-gradient rule on the matched-reference pools. ImageNet Top-1 (\%) at 12,474 updates. $\Delta$ is FGC minus Local (pp).}\label{tab:app-local-adapt}
\begin{tabular}{llrrrrr}\toprule
Model & Rule & Local 17/29 & FGC 17/29 & Local mean & FGC mean & $\Delta$\\\midrule
DeiT-S & Default & 67.932/67.804 & 68.564/68.618 & 67.868 & 68.591 & +0.723\\
DeiT-S & $100\times$ & 68.812/68.756 & 69.616/69.572 & 68.784 & 69.594 & +0.810\\
ViT-S & Default & 67.364/67.420 & 68.328/67.974 & 67.392 & 68.151 & +0.759\\
ViT-S & $100\times$ & 68.814/68.818 & 68.522/68.570 & 68.816 & 68.546 & -0.270\\
\bottomrule\end{tabular}
\end{table}

\paragraph{Earlier recovery checkpoints.}
\Cref{tab:app-checkpoints} adds the block-readout arm to the default-schedule comparison in \cref{fig:recovery-checkpoints}. All rows are measured checkpoints along the original schedule. At 6,174 updates, operator FGC exceeds the 12,474-update MSE endpoint on both backbones in both recovery seeds.

\begin{table}[!htbp]\centering\small
\caption{Default recovery trajectory on the primary 1K pools. Two-seed mean ImageNet Top-1 (\%).}\label{tab:app-checkpoints}
\begin{tabular}{lrrr}\toprule
Preparation & Updates & DeiT-S & ViT-S\\\midrule
Tensor MSE & 3150 & 66.723 & 66.345\\
Tensor MSE & 6174 & 67.206 & 67.122\\
Tensor MSE & 12474 & 66.992 & 67.299\\
FGC: block readout & 3150 & 68.261 & 67.004\\
FGC: block readout & 6174 & 68.643 & 67.886\\
FGC: block readout & 12474 & 68.618 & 67.825\\
FGC & 3150 & 68.141 & 67.019\\
FGC & 6174 & 68.642 & 67.637\\
FGC & 12474 & 68.911 & 67.892\\
\bottomrule
\end{tabular}

\end{table}

\FloatBarrier

\end{document}